\pdfoutput=1

\documentclass[11pt]{article}

\usepackage{ACL2023}

\usepackage{times}
\usepackage{latexsym}

\usepackage[T1]{fontenc}

\usepackage[utf8]{inputenc}

\usepackage{microtype}

\usepackage{inconsolata}

\usepackage{makecell} 

\usepackage{booktabs} 
\usepackage{array}
\usepackage{multirow}
 \usepackage{amsmath}  
\usepackage{hyperref}  
\usepackage{stfloats}  
\usepackage{microtype}
\usepackage{amssymb}
\usepackage{lineno}
\usepackage{graphicx}
\usepackage{times}
\usepackage{latexsym}
\usepackage{tabu}
\usepackage{color}
\usepackage{caption}
\usepackage{setspace}
\usepackage{amsmath}
\usepackage{lineno}
\usepackage{soul}
\usepackage{enumitem}  
\usepackage{bbding}  
\usepackage{CJKutf8}
\usepackage[T1]{fontenc}
\usepackage[utf8]{inputenc}

\title{Facilitating Fine-grained Detection of Chinese Toxic Language: \\ Hierarchical Taxonomy, Resources, and Benchmarks}

\author{Junyu Lu, Bo Xu, Xiaokun Zhang, Changrong Min,  Liang Yang, Hongfei Lin\\
        School of Computer Science and Technology, Dalian University of Technology, China \\ 
        \texttt{(dutljy,kun,11909060)@mail.dlut.edu.cn,(xubo,liang,hflin)@dlut.edu.cn}}

\begin{document}
\begin{CJK*}{UTF8}{gbsn}

\maketitle

\begin{abstract}
\textit{\textbf{Disclaimer}: The samples presented by this paper may be considered offensive or vulgar.}

The widespread dissemination of toxic online posts is increasingly damaging to society. However, research on detecting toxic language in Chinese has lagged significantly. Existing datasets lack fine-grained annotation of toxic types and expressions, and ignore the samples with indirect toxicity. In addition, it is crucial to introduce lexical knowledge to detect the toxicity of posts, which has been a challenge for researchers. In this paper, we facilitate the fine-grained detection of Chinese toxic language. First, we built \textsc{Monitor Toxic Frame}, a hierarchical taxonomy to analyze toxic types and expressions. Then, a fine-grained dataset \textsc{ToxiCN} is presented, including both direct and indirect toxic samples. We also build an insult lexicon containing implicit profanity and propose Toxic Knowledge Enhancement (TKE) as a benchmark, incorporating the lexical feature to detect toxic language. In the experimental stage, we demonstrate the effectiveness of TKE. After that, a systematic quantitative and qualitative analysis of the findings is given. 
\footnote{Resources and codes of this paper are available at \url{https://github.com/DUT-lujunyu/ToxiCN.}}  

\end{abstract}


\section{Introduction}

More and more people have acquired information from social media platforms where posts containing toxic language are also rampant. 
Toxic language is viewed as a rude, disrespectful, or unreasonable comment that is likely to make someone leave a discussion \cite{DBLP:conf/aies/DixonLSTV18}.
Due to its negative impact on individuals and society, toxic language has been rapidly recognized as an increasing concern \cite{DBLP:conf/icwsm/SilvaMCBW16}. 
Recently, researchers have tackled the problem of toxic language detection using the techniques of natural language processing, making great progress in many languages. \cite{DBLP:conf/cikm/MouYL20, DBLP:conf/coling/CaoL20, DBLP:conf/acl/TekirougluCG20, DBLP:conf/icwsm/FountaDCLBSVSK18, zhou2021hate, DBLP:conf/aaai/MathewSYBG021, DBLP:journals/corr/abs-2010-12472, Detoxify}.  

In contrast, the relevant research on Chinese toxic language detection has lagged significantly
\cite{jahan2021systematic}. There are two key issues that have been overlooked. First, existing studies \cite{deng-etal-2022-cold, jiang2022swsr, zhou-etal-2022-towards-identifying} lack a fine-grained annotation of textual toxic types, resulting in hate speech being conflated with general offensive language. Compared to hate speech, general offensive language does not attack groups with special social attributes, and it is just used to emphasize emotions in many cases \cite{DBLP:conf/cscw/0002CTS14}. Like Exp. 1 in Table \ref{introduction}, the insult "\textit{fuck}" can be considered as a modal particle to express surprise. Since there is no equivalence between general offensive language and hate speech, it is crucial to determine their boundary conditions \cite{DBLP:conf/icwsm/DavidsonWMW17}. 

\begin{table*}[htpb]
  \centering
  \small
    \begin{tabular}{m{0.75cm}m{7.5cm}m{1.75cm}m{2.25cm}m{1.5cm}}
    \toprule
    \textbf{Exp .} & \textbf{Post}  & \textbf{Toxic Type} & \textbf{Targeted Group} & \textbf{Expression} \\
    \midrule
    \multirow{2}[2]{*}[3pt]{1} & 我靠！我们居然输了。 & \multirow{2}[2]{*}[3pt]{Offensive} & \multirow{2}[2]{*}[3pt]{-} & \multirow{2}[2]{*}[3pt]{Explicitness} \\
    & \textit{What the fuck! I can't believe we lost!} &       &       &  \\
    \midrule
    \multirow{2}[2]{*}[3pt]{2} &我一看老黑就想吐。 & \multirow{2}[2]{*}[3pt]{Hate} & \multirow{2}[2]{*}[3pt]{Racism} & \multirow{2}[2]{*}[3pt]{Explicitness} \\
    & \textit{I feel like throwing up when I look at n*ggas.} &       &       &  \\
    \midrule
    \multirow{2}[2]{*}[3pt]{3} & 小仙女的事你少管。 & \multirow{2}[2]{*}[3pt]{Hate} & \multirow{2}[2]{*}[3pt]{Sexism} & \multirow{2}[2]{*}[3pt]{Implicitness} \\
    & \textit{Keep your nose out of the fairy's business.} &       &       &  \\  
    \midrule
    \multirow{2}[2]{*}[3pt]{4} & 我的朋友说河南人经常偷井盖。 & \multirow{2}[2]{*}[3pt]{Hate} & \multirow{2}[2]{*}[3pt]{Regional Bias} & \multirow{2}[2]{*}[3pt]{Reporting} \\
    & \textit{My friend said Henan people often steal manhole covers.}   &       &       &  \\
    \bottomrule
    \end{tabular}%
    \caption{Different categories of toxic comment illustration, including \textit{general offensive language} and each hate expression (\textit{explicitness}, \textit{implicitness} and \textit{reporting}).}
  \label{introduction}%
\end{table*}%

In addition, most studies on toxic Chinese language only concentrate on detecting direct and explicit bias and offense. And they lose sight of indirect expressions including implicit hatred (e.g., stereotype and irony) \cite{DBLP:conf/emnlp/ElSheriefZMASCY21} and reporting experiences of discrimination \cite{DBLP:conf/lrec/ChirilMBMOC20}. Due to the absence of direct swear words, these indirect toxic samples are obviously harder to be filtered \cite{DBLP:conf/emnlp/ElSheriefZMASCY21}. To further illustrate the distinction of several expressions, a few examples are listed in Table \ref{introduction}.  Meanwhile, compared to English, Chinese has richer variants of profanity with implicit toxic meaning \cite{zhang2011, DBLP:conf/icdm/SohnL19}, which brings challenge to research on toxic language detection. However, the existing insult lexicon fails to cover these terms. An example is "\textit{fairy}" in Exp. 3, which itself is a positive word and is used here to implicitly attack women. Due to the significance of lexical knowledge to detect toxic language \cite{DBLP:conf/naacl/WiegandRE21, DBLP:conf/acl/HartvigsenGPSRK22}, it is important to construct an insult lexicon containing implicit toxic terms. 

To fill these gaps, we facilitate fine-grained detection of Chinese toxic language. To distinguish hate speech from general offensive language and analyze the expressions of samples, we first introduce \textsc{Monitor Toxic Frame}, a hierarchical taxonomy. Based on the taxonomy, the posts are progressively divided into diverse granularities as follows: \textbf{(I) Whether Toxic, (II) Toxic Type (general offensive language or hate speech), (III) Targeted Group, (IV) Expression Category (explicitness, implicitness, or reporting)}. After taking several measures to alleviate the bias of annotators, we then conduct a fine-grained annotation of posts, including both direct and indirect toxic samples. And \textsc{ToxiCN} dataset is presented, which has 12k comments containing \textit{sexism}, \textit{racism}, \textit{regional bias}, and \textit{anti-LGBTQ}.  



For the convenient detection of toxic language, we construct an insult lexicon against different targeted groups. It contains not only explicit profanities but also implicit words with toxic meanings, such as ironic metaphors (e.g., "\textit{fairy}"). To exploit the lexical feature, we further present a migratable benchmark of Toxic Knowledge Enhancement (TKE), enriching the text representation. In the evaluation phase, several benchmarks with TKE are utilized to detect toxic language, demonstrating its effectiveness. After that, we analyze the experimental results in detail and offer our suggestions for identifying toxic language. The main contributions of this work are summarized as follows:

\begin{itemize}
    \item We present a hierarchical taxonomy, \textsc{Monitor Toxic Frame}, to progressively explore the toxic types and expressions of samples from diverse granularities.
    \item Based on the taxonomy, we propose \textsc{ToxiCN}, a fine-grained dataset of Chinese toxic language. It divides hate speech from offensive language, including samples with not only direct offense but also indirect expressions.   
    \item We present an insult lexicon, and a Toxic Knowledge Enhancement benchmark to incorporate the lexical feature. We evaluate its performance at different levels and conduct an exhaustive analysis.    
\end{itemize}

\section{Related Work}

\begin{table*}[htpb]
\renewcommand{\arraystretch}{1.2}
\center
\small
\begin{tabular}{m{2.5cm}m{1.75cm}m{1.9cm}m{0.85cm}m{1cm}m{1.2cm}<{\centering}m{1.5cm}<{\centering}m{1.5cm}<{\centering}} 
\bottomrule 
\textbf{Work} & \textbf{Source} & \textbf{Scope} & \textbf{Size} & \textbf{Balance} & \textbf{Toxic Type} & \textbf{Expression Category} & \textbf{Implicit Profanity}\\  
\hline 
COLD \cite{deng-etal-2022-cold} & Zhihu, Weibo  & Offensive & 37,480 & 48.1\% & \Checkmark &  & \\ 
\hline 
SWSR \cite{jiang2022swsr} & Weibo  & Hate speech & 8,969 & 34.5\%  &  & \Checkmark   & \\ 
\hline 
CDial-Bias-Utt \cite{zhou-etal-2022-towards-identifying} & Zhihu  &  Hate speech & 13,394 & 18.9\%  &  &  & \\ 
\hline 
CDial-Bias-Ctx \cite{zhou-etal-2022-towards-identifying} & Zhihu  &  Hate speech & 15,013 & 25.9\% &  &  & \\ 
\hline 
\textsc{ToxiCN} (ours) & Zhihu, Tieba  & Offensive and hate speech & 12,011 & 53.8\% & \Checkmark &\Checkmark  & \Checkmark \\ 

\toprule
\end{tabular}  
    \caption{Summary of Simplified Chinese toxic language datasets in terms of \textit{Source}, \textit{Scope}, \textit{Size}, toxic class ratio (\textit{Balance}), and the inclusion of \textit{Toxic Type}, \textit{Expression category}, and the construction of the lexicon containing \textit{Implicit Profanity}.}
\label{dataset_work}
\end{table*}

\textbf{Toxic Language Detection.} Toxic language detection is a high-profile task in the field of natural language processing. Recently, most researchers have utilized methods of deep learning based on the pre-trained language model to tackle this problem \cite{DBLP:conf/cikm/MouYL20, DBLP:conf/coling/CaoL20, DBLP:conf/acl/TekirougluCG20, DBLP:conf/aaai/MathewSYBG021, zhou2021hate}. Two re-trained BERT \cite{DBLP:conf/naacl/DevlinCLT19}, HateBERT \cite{DBLP:journals/corr/abs-2010-12472} and ToxicBERT \cite{Detoxify}, have been specifically proposed to detect toxic language. \citet{DBLP:conf/icwsm/DavidsonWMW17, DBLP:conf/icwsm/FountaDCLBSVSK18, DBLP:conf/aaai/MathewSYBG021} attempted to distinguish hate speech from offensive language, while \citet{DBLP:conf/emnlp/ElSheriefZMASCY21, DBLP:conf/acl/HartvigsenGPSRK22} explored the benchmark of implicit and latent hate speech and \citet{DBLP:conf/lrec/ChirilMBMOC20, DBLP:conf/coling/Perez-Almendros20} considered testing for reporting related to hate speech and behavior. Meanwhile, for the construction of the toxic language dataset, there have been some studies focused on how to improve the reliability of the annotation process to alleviate the subjective bias of annotators \cite{DBLP:conf/naacl/WaseemH16, DBLP:conf/acl/ZeinertID20}.

\textbf{Linguistic Research of Chinese Toxic Language.} Chinese toxic language has been researched extensively in language studies and sociolinguistics. \citet{zhou2000} analyzed Chinese vernacular novels and summarized the common rhetorical methods of offensive language. According to \citet{wang2009abusive, li2020swearwords}, insults are more easily expressed through variants. Due to the lack of morphological markers, authors often make sentences based on language sense and consensus \cite{zhang2004}, expressing hatred more concisely compared to Indo-European \cite{zhang2005}. \citet{zhang2011, DBLP:conf/icdm/SohnL19} compared insults in English and Chinese and discovered that Chinese has a richer variety of profanity due to its unique culture and linguistics. 
These linguistic features bring challenges to Chinese toxic detection.
Recently, some Chinese toxic language datasets have been constructed \cite{deng-etal-2022-cold, jiang2022swsr, zhou-etal-2022-towards-identifying}. However, they fails to separate hate speech from general offensive language, and overlooks toxic samples containing indirect expressions. Besides that, it lacks the construction of the lexicon containing implicit profanities. In this work, we fill these gaps to facilitate fine-grained detection of Chinese toxic language. Here we list \tablename~\ref{dataset_work} to compare these studies with our \textsc{ToxiCN}. 

%



\section{Dataset Construction}
\subsection{Overview}

In this section, we describe the construction of \textsc{ToxiCN} dataset. We first introduce the process of data collection and filtering. Then, \textsc{Monitor Toxic Frame} is presented as the guideline for labeling. After adopting several measures to mitigate biases in the process of labeling, we implement a hierarchical fine-grained annotation based on the frame. The Inter-Annotator Agreement (IAA) of each individual granularity is explored. Finally, related statistics of \textsc{ToxiCN} are shown.

 %


\subsection{Data Collection and Filtering}
To avoid a high degree of homogenization of data, we crawl the published posts from two public online media platforms, \textit{Zhihu} and \textit{Tieba}. Both platforms are representative of local users in China and have active communities about specific topics. Due to the filtering mechanism of the websites, the proportion of posts containing toxic language is relatively sparse \cite{deng-etal-2022-cold}. Thus, we first limit the scope of crawled data under several sensitive topics, including \textit{gender}, \textit{race}, \textit{region}, and \textit{LGBTQ}, which are easily debated on the Internet. And then, we list some keywords for each topic and utilize them to extract a total of 15,442 comments without replies. We remove the samples where the text is too brief to have actual semantics, such as phrases consisting of only inflections and auxiliaries. And dirty data is also deleted, including duplicated samples and irrelevant advertisements. In the end, 12,011 comments are retained\footnote{We also attempted to crawl posts from \textit{Weibo} referenced \cite{deng-etal-2022-cold}, however, due to the filtering mechanism, there is no guarantee that sufficient samples will be obtained under each topic, such as "\textit{race}" and "\textit{LGBTQ}", resulting in a relatively homogeneous crawl. Therefore, we finally choose \textit{Zhihu} and \textit{Tieba} as our data sources.}.

In the stage of data cleaning, we focus on normalizing the unique form of expression in web text adopted from \citet{DBLP:conf/semeval/AhnSPS20}. The extra newline characters and spaces in the original text are also deleted. To prevent privacy leakage, we desensitize the data, filtering out \textit{@USERs}, attached links, and pictures. Due to the possibility of carrying important emotional information \cite{DBLP:conf/aaai/MathewSYBG021}, we reserve emojis for toxic detection. 


\subsection{Data Annotation}

\subsubsection{Monitor Toxic Frame}
To determine the downstream subtasks and establish the finalized guidelines for labeling, a standardized taxonomy is needed. Here we build a hierarchical taxonomy called \textsc{Monitor Toxic Frame} (in \figurename~\ref{frame_exp}). It has three levels including four diverse aspects to identify toxic content and further in-depth analysis. The specific implementations are as follows:


\begin{figure}[htpb]
\centering
\includegraphics[width=7.5cm]{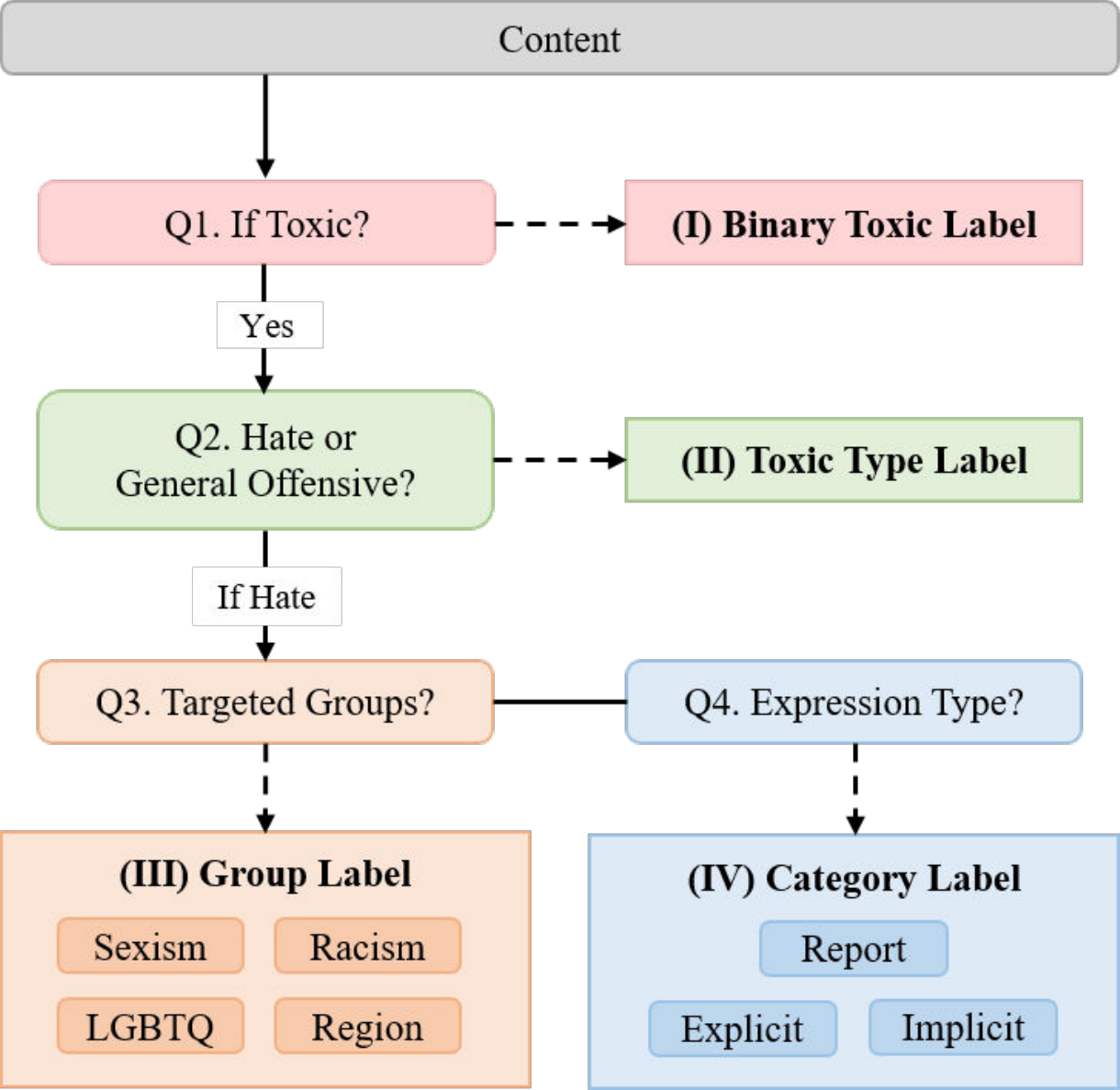}
\caption{\textsc{Monitor Toxic Frame} illustration. The framework introduces four questions to determine whether a comment is general offensive language or hate speech, and further analyzes the attacked group and expression type.}
\label{frame_exp}
\end{figure}

\textbf{Toxic Identification}. The first level of our framework is a binary annotation to determine whether the comment contains toxic language, which is the foundation of subsequent labeling. 
In this work, general offensive language and hate speech are highlighted. 

\textbf{Toxic Type Discrimination}. The second level is to distinguish general offensive language and hate speech. Based on \citet{DBLP:conf/naacl/WaseemH16} and \citet{DBLP:journals/csur/FortunaN18}, we list several criteria for identification of hate speech: 1) attacking specific groups, or 2) inciting others to hate minorities, or 3) creating prejudice, rejection, or disgust for minorities based on stereotypes and distorted facts, or 4) using sarcasm or humor to ridicule groups, despite the publisher may not be malicious. In contrast, general offensive language is not insulting to targets with specific social attributes \cite{DBLP:conf/icwsm/DavidsonWMW17}.   

\textbf{Targeted Group and Expression Type Detection}. In the third level, we further explore the targeted group and expression type of toxic language. If the content contains hate speech, its characteristics of the target group are specified, including \textit{sexism}, \textit{racism}, \textit{regional bias}, and \textit{anti-LGBTQ}. Since multi-class groups may be attacked in a text, this task is categorized as multi-label classification.
Meanwhile, we determine the categories of toxic expressions, containing \textit{explicitness},  \textit{implicitness}, and \textit{reporting}. In these expressions, 1) explicitness is obviously offensive, inciting, or prejudiced against minority groups, and 2) implicitness refers to the subtle or even humor expression without strong exclusion, such as microaggressions and stereotyping \cite{DBLP:conf/acl/HartvigsenGPSRK22}, and 3) reporting only documents or condemns experience of attack and discrimination against minorities, and the publisher does not express hatred and prejudice \cite{DBLP:conf/lrec/ChirilMBMOC20}. As the expression category of general offensive language is necessarily explicit, we focused on the expressions in hate speech. This granularity is set as multi-classification.

\subsubsection{Mitigating Bias}
The subjective biases of the annotators negatively impact the quality of the dataset \cite{DBLP:conf/naacl/WaseemH16}. Therefore, it is significant to mitigate these biases during the design and construction of annotations. For this purpose, we adopt the following measures:  We first guarantee the diversity of annotators in terms of background information, including gender, age, race, region, and study. All participants major in linguistics and have been systematically trained. The demographics of annotators are shown in \tablename~\ref{demographics}. Then, we make a progressive analysis of the toxic content contained in the crawled posts, and initially determine the labeling rules for various granularities. After a couple of iterations of small-scale annotation tests and discussions of edge cases, the final criteria are established.


\begin{table}[htpb]
\renewcommand{\arraystretch}{1.2}
\center
\small
\begin{tabular}{m{2.5cm}<{\centering}m{4cm}<{\centering}} 
\bottomrule 
\textbf{Characteristic} & \textbf{Demographics}\\  
\hline 
Gender & 5 male, 4 female    \\ 
Age & 5 age < 25, 4 age $\geq $ 25 \\  
Race & 6 Asian, 3 others  \\
Region &  From 5 different provinces \\
Education & 2 BD, 4 MD, 3 Ph.D. \\
\toprule
\end{tabular}
\caption{Annotators demographics.}
\label{demographics}%
\end{table}

\begin{table*}
\renewcommand{\arraystretch}{1.2}
\small
  \centering
    \begin{tabular}{m{1.5cm}<{\centering}|m{1.1cm}<{\centering}|m{1.1cm}<{\centering}|m{1.1cm}<{\centering}|m{1.1cm}<{\centering}|m{1.1cm}<{\centering}m{1.1cm}<{\centering}m{1.1cm}<{\centering}|m{1.1cm}<{\centering}||m{1.1cm}<{\centering}}
    \bottomrule
    \multirow{2}[4]{*}[5pt]{\textbf{Topic}} & \multirow{2}[4]{*}[5pt]{\textbf{N-Tox.}}  & \multirow{2}[4]{*}[5pt]{\textbf{Tox.}} & \multicolumn{5}{c|}{\textbf{Toxic Category}} & \multirow{2}[4]{*}[5pt]{\textbf{Total}} & \multirow{2}[4]{*}[5pt]{\textbf{Avg. \textit{L}}} \\
\cline{4-8}     &      &       & \textbf{Off.} & \textbf{Hate} & \textbf{H-exp.} & \textbf{H-imp.} & \textbf{H-rep.} &       &         \\ 
    \hline
    \textbf{Gender} &1,805    & 2,153      &316       &1,837       &1,055       &693       &89       &3,958       &35.26  \\
    \textbf{Race} &1,602       &2,084       &229       &1,855       &1,041       & 711      & 103      &3,686       &36.93  \\
    \textbf{Region} & 1,222 &  1,148     &  82     &  1,066     & 172      & 292      & 602 &2,370   & 40.26 \\
    \textbf{LGBTQ} & 921      &1,076       &189       &887       &469       &299       &119       &1,997       &44.96  \\
    \hline
    \textbf{Total} & 5,550   &6,461    & 816      & 5,645      & 2,737      & 1,995      & 913      & 12,011      & 38.37 \\
    \toprule
    \end{tabular}%
    \vspace{-0.05in}
\caption{Basic statistics of \textsc{ToxiCN}, listing the number of non-toxic (\textit{N-Tox.}) and toxic (\textit{Tox.}) comments, containing general offensive language (\textit{Off.}) and each hate expression categories (including explicitness (\textit{H-exp.}), implicitness (\textit{H-imp.}) and reporting (\textit{H-rep.})). And \textit{Avg. \textit{L}} is the average length of samples.}
\vspace{-0.05in}
\label{statistics}
\end{table*}%

\subsubsection{Annotation Procedure}
The annotation procedure consists of two steps: pseudo labeling and main manual annotation. Meanwhile, the initial construction of the insult lexicon is implemented.  

\textbf{Pseudo Labelling.} To reduce the burden of manual annotation, we retrieve the samples containing insults, most of which are obviously toxic. Specifically, we first build an original lexicon of explicit profanity words, integrating two existing profanity resources, including HateBase\footnote{https://hatebase.org/}, the world's largest collaborative and regionalized repository of multilingual hate speech, and SexHateLex\footnote{https://zenodo.org/record/4773875} \cite{jiang2022swsr}, a large Chinese sexism lexicon. Then, an iterative approach is employed to match out profanity-laced comments using regular expressions. The swearwords contained in these samples that are not in the lexicon are further collected. We assign a toxic pseudo-label for each sentence containing insults. After several iterations, the remaining samples are directly pseudo-labeled as non-toxic. The statistics illustrate that this method is simple and effective, correctly separating about 50\% of toxic samples from ToxiCN. See Table \ref{if_w_insult} from Appendix \ref{Appendix_sta} for a more detailed report. 

\textbf{Main Manual Annotation.} Based on \textsc{Monitor Toxic Frame}, we implement the main annotation of \textsc{ToxiCN}. Most samples pseudo-labeled as toxic are directly categorized as general offensive language or explicit toxic language. Afterwards, due to the low frequency variants of insults and implicit toxicity expressions, the remaining pseudo-labeled as non-toxic samples have to be re-annotated in a hierarchical manner. Meanwhile, the implicit insults contained in these samples are added to the previous profanity list. We utilize the open source text annotation tool Doccano\footnote{https://github.com/doccano/doccano} to facilitate the labeling process. Each comment is labeled by at least three annotators and a majority vote is then used to determine the final label. After annotation, we explore the Inter-Annotator Agreement (IAA) of \textsc{ToxiCN}, and Fleiss’ Kappa of each hierarchy is shown as \tablename~\ref{IAA}.

\begin{table}
\renewcommand{\arraystretch}{1.2}
\center
\small
\begin{tabular}{m{1.45cm}<{\centering}m{1.45cm}<{\centering}m{1.45cm}<{\centering}m{1.45cm}<{\centering}} 
\bottomrule 
\textbf{If Toxic} & \textbf{Toxic Type}  & \textbf{Targeted} &  \textbf{Expression}  \\  
\hline
0.62 & 0.75  & 0.65 & 0.68  \\  
\toprule
\end{tabular} 
\vspace{-0.05in}
    \caption{Fleiss’ Kappa for different granularities.}
    \vspace{-0.05in}
\label{IAA}
\end{table}

\subsection{Data Description} 

In the data analysis phase, we first describe the dataset from the topic of comments. The basic statistics of \textsc{ToxiCN} are shown in \tablename~\ref{statistics}. We note that there is a sample imbalance between different categories of toxic samples. Specifically, the \textit{Off.} class represents only 6.8\% of the overall dataset. Because the data distribution reflects the true situation of the platforms \cite{DBLP:conf/aaai/MathewSYBG021}, we do not apply additional treatment to the imbalance. In addition, since a single case from a topic may attack multi-class groups, we further record the sample size of hate speech against various target categories. From \tablename~\ref{target2expression}, we can see that the distribution of expressions is different for each group. For example, most samples containing regional bias are reporting, which is uncommon in other categories. More statistical details are listed in Appendix \ref{Appendix_sta}.

\begin{table}[htpb]
\renewcommand{\arraystretch}{1.2}
\small
  \centering
    \begin{tabular}{m{1.5cm}<{\centering}|m{1cm}<{\centering}m{1cm}<{\centering}m{1cm}<{\centering}||m{1cm}<{\centering}}
    \bottomrule
    \textbf{Group} & \textbf{H-exp.} & \textbf{H-imp.} & \textbf{H-rep.} & \textbf{Total} \\
    \hline
    \textbf{Sexism} &1,259       &887       &156       &2,302  \\
    \textbf{Racism} &1,149       &660       &65       &1,874  \\
    \textbf{RGN. B.} &295       &384       &610      &1,289  \\
    \textbf{Anti-L+} &671       &383       &121       &1,075  \\
    \toprule
    \end{tabular}%
    \vspace{-0.05in}
\caption{Sample size of various hate expressions for each attacked group label. \textit{RGN. B.} refers to regional bias and \textit{Anti-L+} is short for anti-LGBTQ.}
\vspace{-0.05in}
\label{target2expression}%
\end{table}%



\section{Insult Lexicon}\label{lexicon} 


We divide the insult lexicon built in the process of annotation into five categories according to the attacking object. The lexicon includes sexism, racism, regional bias, anti-LBGTQ, and general swearwords, referring to the swear words that can be used to offend any group. The final dictionary contains 1,032 terms, including not only explicit vulgar words but also implicit insults. In addition, as the Internet generates a wealth of new insults every year, it is vital to explore their origins. Therefore, for the convenience of the follow-up research, we further analyze the rules for the derivation of Internet swear words from the proposed insult lexicon. Specifically, we briefly summarize them in terms of both surface features and their actual meaning.  More related terminology notes and examples of profanity are illustrated in Appendix \ref{Appendix_rules}. 

\textbf{Surface Features.} To circumvent the filtering mechanism, netizens change the original insults and create new variants with similarities in glyphs and pronunciation \cite{chen2012, zhang2011}, which are called \textit{deformation} and \textit{homophonic}, respectively. In addition, Chinese characters are at times replaced with other language codes in some profanities \cite{li2020swearwords}, creating \textit{code-mixing words} or \textit{abbreviations}.  

\textbf{Actual Meaning.} Internet users often introduce implicit terms to attack or defame the target groups, including usage of \textit{metaphor} and \textit{irony} \cite{chen2012}. Besides that, some \textit{borrowed words} also contain specific prejudices, which are used in implicit toxic comments \cite{shi2010web}. Compared to variants based on surface features, these terms with deep semantics have to be detected with background knowledge.  

\section{Methodology}
\begin{figure}
\centering
\includegraphics[width=7.5cm]{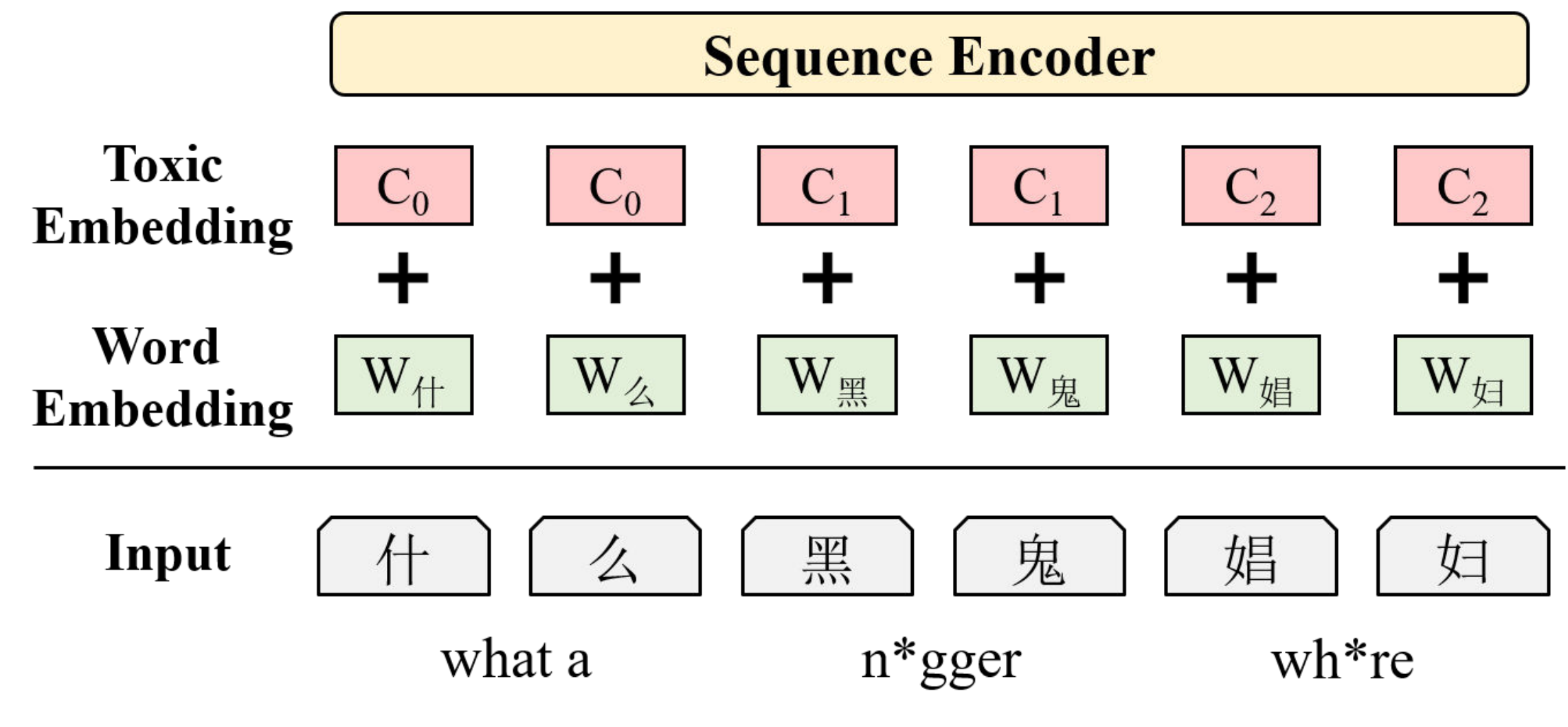}
\caption{Toxic Knowledge Enhancement (TKE) illustration. Here we set the category representations of non-toxic terms, racist terms, and sexist terms as $C_0$, $C_1$, and $C_2$, respectively.}
\label{tke}
\end{figure}

In view of the significance of lexical knowledge to detect toxic language, we propose a migratable benchmark of Toxic Knowledge Enhancement (\textbf{TKE}), incorporating the lexical feature to enrich the original sentence representation. The illustration of TKE is shown in \figurename~\ref{tke}. 

\begin{table*}[htbp]
\renewcommand{\arraystretch}{1.2}
\small
  \centering
    \begin{tabular}{m{1.3cm}|m{0.75cm}<{\centering}m{0.75cm}<{\centering}m{0.75cm}<{\centering}|m{0.75cm}<{\centering}m{0.75cm}<{\centering}m{0.75cm}<{\centering}|m{0.75cm}<{\centering}m{0.75cm}<{\centering}m{0.75cm}<{\centering}|m{0.75cm}<{\centering}m{0.75cm}<{\centering}m{0.75cm}<{\centering}}
    \bottomrule
    \multirow{2}[4]{*}{} & \multicolumn{3}{c|}{\textbf{Toxic Identification}} & \multicolumn{3}{c|}{\textbf{Toxic Type}} & \multicolumn{3}{c|}{\textbf{Targeted Group}} & \multicolumn{3}{c}{\textbf{Expression Category}} \\
\cline{2-13}          & $P$     & $R$     & $F_1$    & $P$     & $R$     & $F_1$    & $P$     & $R$     & $F_1$    & $P$     & $R$     & $F_1$ \\
    \hline
    BTC  &  $64.2$  &  $ 53.0$    & $45.9$      & -      &  -     &   -    &  -     & -      &  -     &   -    &  -     & - \\
    \hline
    BiLSTM  & $73.7_{0.6}$  & $72.7_{0.4}$      & $72.9_{0.4}$      &$77.7_{2.2}$   &$70.4_{1.2}$  & $73.7_{0.5}$ & $61.1_{1.2} $    & $64.4_{0.7}$   &$ 62.2_{0.5}$    &  $49.7_{1.6}$     &  $48.6_{1.1} $    &$ 48.0_{1.0}$   \\
    BiLSTM*  & $75.4_{0.5}$  &$ 74.8_{0.5}$  & $74.9_{0.4}$        & $79.2_{2.0} $     & $69.6_{1.4} $     & $73.6_{0.4}$      &  $58.8_{0.9}$     & $68.6_{1.3} $     & $62.8_{0.7} $     &  $51.2_{1.9} $    &  $56.8_{1.7}$     & $53.5_{1.3}$ \\
    BERT  & $80.0_{0.2}$     & $79.7_{0.2}$      & $79.7_{0.2}$      & $\textbf{82.8}_{0.9} $     & $73.1_{1.0}$      & $77.3_{0.4}$      & $71.1_{0.9}$      & $71.9_{1.0}$      & $72.2_{0.4}$      & $\textbf{55.9}_{1.1}$      & $56.4_{1.3}$      & $55.3_{1.1}$  \\
    BERT*  & $80.4_{0.3}$      & $80.2_{0.3}$      & $80.0_{0.3}$      & $80.2_{1.3}$      & $75.2_{0.9}$      & $77.3_{0.2}$      & $\textbf{73.3}_{1.2}$      & $72.6_{0.9}$      & $72.6_{0.4}$       & $53.5_{2.0}$      & $60.9_{0.5}$      & $55.9_{0.9}$ \\
    RoBERTa & $80.8_{0.2}$      & $80.2_{0.3}$      & $80.3_{0.3}$      & $80.5_{0.9}$      & $74.6_{0.5}$      & $77.3_{0.3}$      & $71.8_{1.4}$      & $73.9_{1.1}$      &  $72.6_{0.5}$          & $54.2_{1.2}$      & $58.7_{1.1}$      & $55.8_{0.6}$ \\
    RoBERTa* & $\textbf{80.9}_{0.3}$       & $\textbf{80.5}_{0.3}$      & $\textbf{80.6}_{0.3}$      & $79.8_{1.8}$      & $\textbf{76.1}_{1.2}$      & $\textbf{77.7}_{0.4}$      &$ 72.5_{0.9}$      & $\textbf{74.0}_{1.0}$      & $\textbf{73.0}_{0.5}$       & $54.4_{1.6}$      & $\textbf{61.3}_{1.2}$      & $\textbf{56.8}_{0.4}$ \\
    \toprule
    \end{tabular}%
    \vspace{-0.025in}
    \caption{Evaluation of each subtask. Results show the mean and s.d. (subscript) of $P$, $R$, and $F_1$, where BTC denotes Baidu Text Censor, \textit{$^*$} refers to the introduction of TKE to the baseline, and the \textbf{bold} score represents the best obtained values. Because BTC is an online API with no training required, we use it to perform the toxic identification of all the samples in \textsc{ToxiCN}.} 
    \vspace{-0.025in}
  \label{result}%
\end{table*}%

For a given sentence $S = \{x_1, x_2, ..., x_n\}$, each token $x_i$ is embedded as $w_i \in \mathbb{R}^d $, which is a vector representation in $d$-dimensional space. Inspired by \citet{zhou2021hate}, we design a toxic embedding to introduce lexical knowledge. Specifically, we first employ the n-gram to determine whether $x_i$ is a subword of an insult, and if so, its attacked group is further indicated. Then, we randomly initialize the group category representation as $C = (c_0, c_1, ..., c_m)$, where $c_i \in \mathbb{R}^d$, $c_0$ refers to non-toxic term and $m$ is the number of categories of the insult lexicon. In this work, $m = 5$. Based on the $c_i$, we further propose the definition of toxic embedding $t_i$ of $x_i$:    

\begin{equation}
t_i=
\begin{cases}
c_o, & \begin{small}\text{if } x_i \text{ is non-toxic.}\end{small}\\
c_j, & \begin{small}\text{if } x_i \text{ is from the } j^{th} \text{ category.} \end{small} 
\end{cases}
\end{equation}

Since the element-wise addition of multiple linear vector representations is an efficient way to fully incorporate various information \cite{DBLP:conf/nips/MikolovSCCD13}, we utilize this method to integrate toxic and word embedding. The enhanced representation of $x_i$ is $w_i' = w_i + \lambda t_i$, where $\lambda \in [0, 1]$ is a weighting coefficient to control the ingestion of toxic knowledge. The ultimate sentence embedding of $S$ is $\{w_1', w_2', ..., w_n'\}$, which is the input of the connected sequence encoder. Due to its convenience, TKE can be migrated to any pre-trained language model (PLM). 

\section{Experiments}

\subsection{Baselines}
Here we introduce the baselines of experiments. Several PLMs are utilized as encoders as follows. And we use a fully-connected layer as the classifier for several subtasks.


\textbf{BiLSTM}. This method employs the word vector of Tencent AI Lab Embedding\footnote{https://ai.tencent.com/ailab/nlp/zh/embedding.html}, a static word vector with 200-dimensional features, and integrates contextual information using BiLSTM. We concatenate the last hidden states from both forward and backward directions to obtain the final sentence embedding.    

\textbf{BERT} \cite{DBLP:conf/naacl/DevlinCLT19} and \textbf{RoBERTa} \cite{DBLP:journals/corr/abs-1907-11692}. The two most commonly used Chinese transformer-based PLMs, \texttt{bert-based-} \texttt{chinese}\footnote{https://huggingface.co/bert-base-chinese}  and \texttt{roberta-base-chinese}\footnote{https://huggingface.co/hfl/chinese-roberta-wwm-ext}, are used as benchmarks. In the experiment, the pooled output of the encoder is utilized as the input of the connected classifier.    



Besides the above deep learning based methods, we also evaluate the performance of \textbf{Baidu Text Censor}\footnote{https://ai.baidu.com/tech/textcensoring}, an online API to identify toxic content. Due to the function limit, we only utilize it in the first subtask of binary toxic identification.

\subsection{Implementation}
We employ the widely used metrics of weighted precision ($P$), recall ($R$), and $F_1$-score ($F_1$) to evaluate the performance of models. Weighted cross entropy is utilized to address the problem of category imbalances, and the optimizer is AdamW. An early stopping mechanism is applied in the training phase. All the samples in \textsc{ToxiCN} are split into a training set and a test set with a ratio of 8:2. We fine-tune the baselines and reserve the best performing models and hyperparameters on the test set, and the same experiments are repeated 5 times by changing the random seeds for error reduction. All experiments are conducted using a GeForce RTX 3090 GPU. More details are shown in Appendix \ref{Appendix_exp}. 


\subsection{Results and Discussions}
In this section, we present our experimental results and progressively analyze the following three questions, respectively:

\textbf{RQ1: Performance of Different Subtasks.}
We evaluate the performance of each baseline and the contribution of TKE at different granularities of toxic language detection. The experimental results are shown in \tablename~\ref{result}. From the results, we can observe that:

(1) Compared with Baidu Text Censor, deep learning based methods achieved better performance. A plausible explanation is the filtering mechanism of the online API mainly depends on the keyword dictionary. Therefore, it cannot effectively detect the toxicity of sentences containing indirect expressions of hate speech. In addition, the performance of the pre-trained language model based on dynamic word representation (e.g., BERT, RoBERTa) is much better than on static embedding. And in these baselines, RoBERTa is the most effective on several subtasks.

(2) Overall, the models introducing TKE have improved the performance of several subtasks, illustrating the effectiveness of representation incorporating toxic lexical knowledge. Among them, TKE leads to the greatest enhancement in the detection of expression category, with an average improvement of 2.7\% of each baseline. This result shows that lexical information can improve the ability of the model to detect toxic language in different expressions. Meanwhile, we also find TKE does not bring significant improvement in toxic type discrimination. This is because many insults are widely contained in both general offensive language and hate speech. Therefore, the introduction of toxic embedding does not distinguish well between these two kinds of speech.   

\begin{figure}[htpb]
\centering
\includegraphics[width=7.5cm]{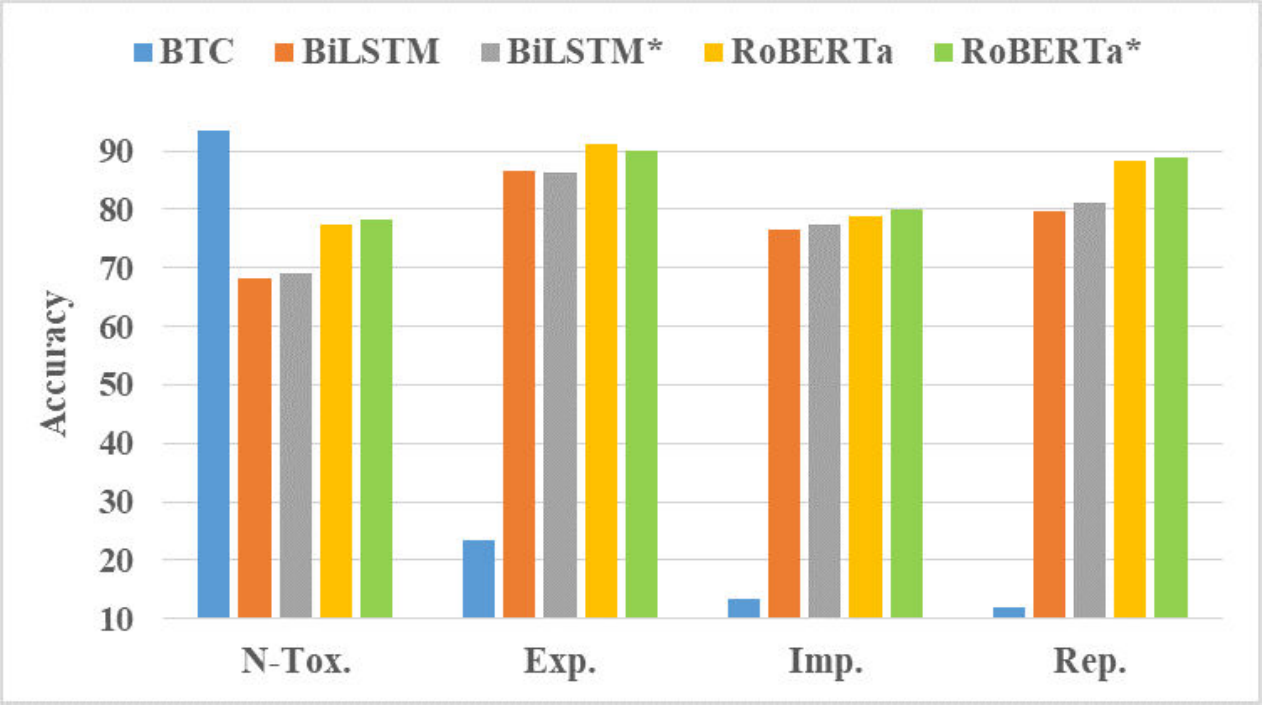}
\caption{Accuracy towards samples with different expressions, containing non-toxicity (\textit{N-Tox.}), explicitness (\textit{Exp.}), implicitness (\textit{Imp.}), and reporting (\textit{Rep.}).}
\label{toxic_small}
\end{figure}

\textbf{RQ2: Detection of Each Toxic Subtype.}
The complementary experiment is conducted to further evaluate the performance of models to identify the toxic content with different expressions. Specifically, we utilize the optimal trained models for the subtask of toxic identification to detect the samples in the test set. After that, we separately analyze the accuracy of sentences labeled as non-toxic and each expression category. The results are shown in \figurename~\ref{toxic_small}.

Based on the result, it is noteworthy that:

(1) Compared to explicit toxic language, the accuracy of implicit expression is significantly lower, with a difference of around 10\%. The reason is that many samples containing implicit toxicity are mistakenly classified as non-toxic due to the absence of explicit insults. 

(2) In spite of more training data, the performance of models to detect implicit toxicity is worse than reporting about 5\%. This is because the reporting contains references to actors such as "\textit {he/she said...}" which support decisions of models. 

(3) The introduction of TKE increases the accuracy of the model for implicit toxicity and reporting samples. It illustrates that the implicit lexcial knowledge enhances the ability of models to detect toxic samples with indirect expressions. 

\textbf{RQ3: Error Analysis.}
For more insight into the performance of TKE, we perform a manual inspection of the set of samples misclassified by all the models. Two main types of errors are summarized. Here we list the following two samples in the set for illustrative purposes:


\begin{description}[noitemsep]
    \item[~\textit{Exp. 1}]\begin{small}
    北京高考400分上清华\end{small} ~---~\textit{Toxic} \\
    \textit{~(It is sufficient to get into THU with a NEMT result of 400 points in Beijing.)} 
    \item[~\textit{Exp. 2}] \begin{small}
    他以前发帖说过自己是~\underline{\textbf{txl}}。 \end{small} ~---~\textit{Non-Toxic} \\
    \textit{~(He has posted before that he is \underline{\textbf{gay}}.)} 
    \vspace{0.05in}
\end{description}

\textbf{Type I error} refers to sentences annotated as \emph{toxic}, but classified as \emph{non-toxic} by the models. This kind of error usually occurs in the detection of samples containing implicit bias, caused by a lack of background information on the semantic level. Like Exp. 1, supported by external knowledge, including 400 is a relatively low score on the NEMT with a full score of 750, and THU is one of the best universities in China, it can be known that the publisher uses a fake message to express implicit regional bias against Beijing.

\textbf{Type II error} denotes to instances labeled as \emph{non-toxic}, while detected as \emph{toxic}. The samples with this error usually contain toxic token-level markers, such as swear words and pronouns of minority groups. The training data with these markers is often labeled as toxic language, leading to spurious associations in the models \cite{DBLP:conf/eacl/ZhouSSCS21}. Like Exp. 2, where "txl" (meaning "\textit{gay}") causes models to make decisions based only on statistical experience rather than incorporating context information. 

From the error analysis, we note that it remains a challenge to integrate richer external knowledge without reducing spurious biases of models. In future work, we will further explore methods of introducing knowledge enhancement for toxic language detection.

\section{Conclusion and Future Work}
 Due to the rampant dissemination of toxic online language, an effective detection mechanism is essential. In this work, we focus on Chinese toxic language detection at various granularities. We first introduce a hierarchical taxonomy \textsc{Monitor Toxic Frame} to analyze the toxic types and expressions. Based on the taxonomy, we then propose a fine-grained annotated dataset \textsc{ToxiCN}, including both direct and indirect toxic samples. Due to the significance of lexical knowledge for the detection of toxic language, we build an insult lexicon and present a benchmark of Toxic Knowledge Enhanced (TKE), enriching the representation of content. The experimental results show the effectiveness of TKE on toxic language detection from different granularities. After an error analysis, we suggest that both knowledge introduction and bias mitigation are essential. We expect our hierarchical taxonomy, resources, benchmarks, and insights to help relevant practitioners in the identification of toxic language.

\section*{Limitations}

Despite the fact that some measures have been implemented to minimize bias in labeling, we are still explicitly aware that our dataset may contain mislabeled data due to differences in the subjective understanding of toxic language by the annotators. Meanwhile, due to the limitation of data coverage, our benchmark is not practical for all types of toxic comments.   

For reasons of intellectual property, we only capture the comments rather than the full text, which affects the actual semantics of the sentence to some extent. Besides that, non-textual features are not taken into account in this work, such as images and meta information about publishers. In future work, we will further research span-level and multi-modal toxic language detection.

\section*{Ethics Statement}

We strictly follow the data use agreements of each public online social platform and double-check to ensure that there is no data relating to user privacy. The opinions and findings contained in the samples of our presented dataset should not be interpreted as representing the views expressed or implied by the authors. We hope that the benefits of our proposed resources outweigh their risks. All resources are for scientific research only.

\section*{Acknowledgment}
This research is supported by the Natural Science  Foundation of China (No. 62076046, 62006034). We would like to thank all reviewers for their constructive comments.

\bibliography{acl2023}
\bibliographystyle{acl_natbib}
\appendix
\renewcommand\thefigure{\Alph{section}\arabic{figure}}    
\renewcommand\thetable{\Alph{section}\arabic{table}} 

\section{Sample}

We adopt JSON file to store \textsc{ToxiCN} dataset, which is a mainstream coding specification to facilitate machine-readable. The construction of data is \textit{Sample = (ID, Platform, Topic, Text, Toxic, Hate, [Group], [Expression])}, where \textit{Toxic} and \textit{Hate} denote whether the sentence contains toxic language or hate speech, respectively. And if it is not biased, \textit{Group} and \textit{Expression} are set to empty. Here we provide two samples in \figurename~\ref{sample}.

\vspace{-0.05in}
\setcounter{figure}{0}    
\begin{figure}[htpb]
\centering
\vspace{-0.1in}
\includegraphics[width=8cm]{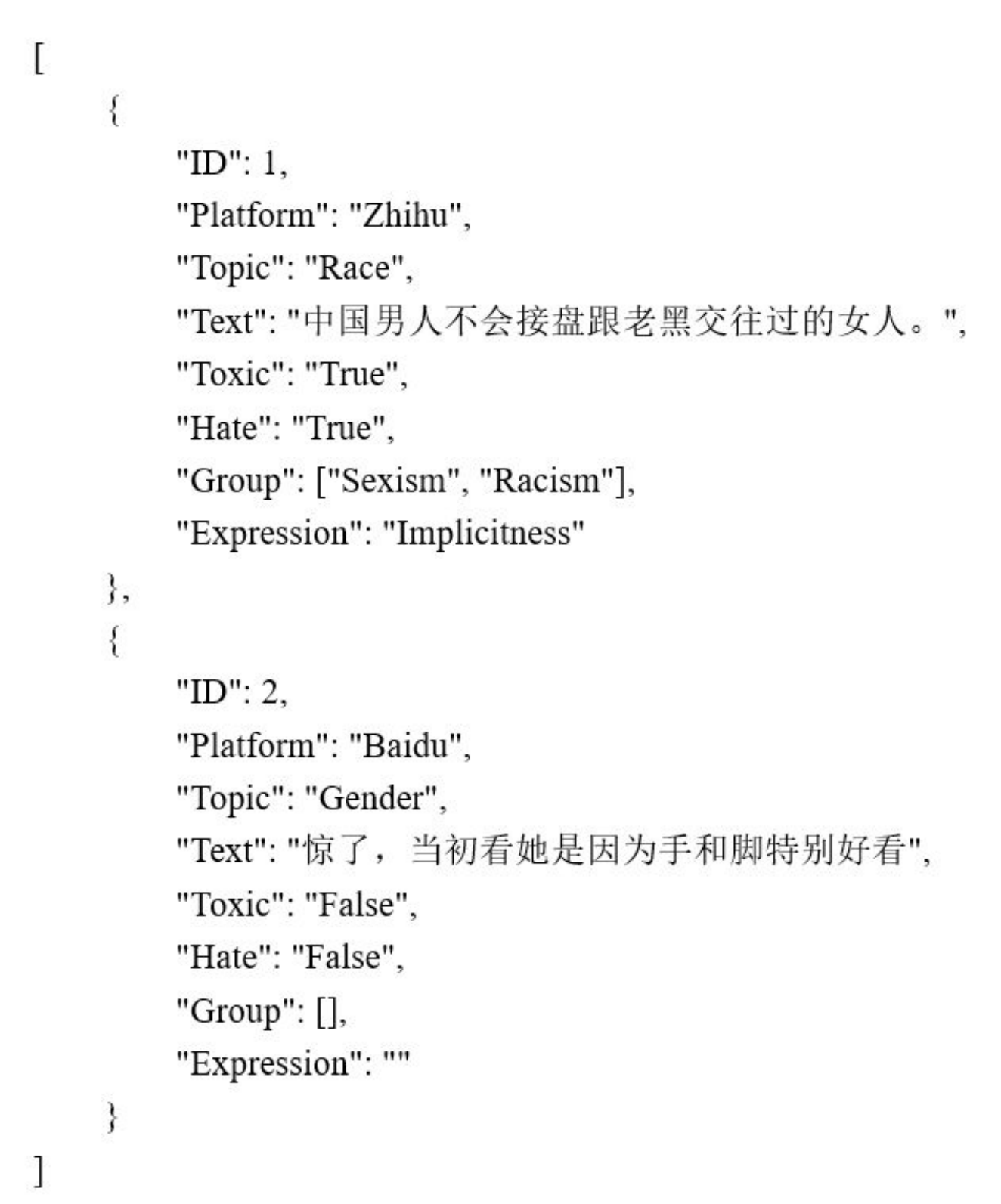}
\vspace{-0.2in}
\caption{Two samples of \textsc{ToxiCN}.}
\label{sample}
\end{figure}

\section{Details of Annotation}
 In the process of annotation, if the root category is wrong, labels of its subcategories, which are annotated by minorities, will be discarded. Then, new raters will be introduced to label this sample. And the Kappa of this sample is recalculated. Based on the IAA shown in \tablename~\ref{IAA}, the most disagreement stems from the phase of “Toxic Identification” and the Kappa is 0.62, caused by implicit hate speech (\textit{H-imp}) containing language techniques like humor. Although we regard these samples as toxic in the rules, some annotators believe that, due to subjective reasons, their toxicity intensity is insufficient to classify them as toxic. And in the “Targeted Group” with a Kappa of 0.65, “\textit{sexism}” and “\textit{anti-LGBTQ}” can also easily be confused.

\section{Derivative Rules of insults}\label{Appendix_rules} 


\setcounter{table}{0}  
\begin{table*}[t]
\renewcommand{\arraystretch}{1.2}
\center
\small
\begin{tabular}{m{3.2cm}m{2.4cm}m{4.1cm}m{2.7cm}m{1.1cm}} 
\bottomrule 
\textbf{Term} & \textbf{Literal Meaning} & \textbf{Composition} &  \textbf{Actual Meaning} & \textbf{Category}  \\  
\hline 
默(mò) & silence  &黑(hēi)~犬(quǎn)~→~black dog & n*gger & racial \\ 
南(nán)~满(mǎn)  & South Manchu &南满~→~南蛮(mán) & southern barbarians & regional  \\   
蠢驴 & silly donkey & - & foolish people &general\\ 
txl & txl & txl~→~同(tóng)~性(xìng)~恋(liàn) & gay
 & anti-L+\\
ni哥(gē)  & ni brother & ni+ger~→~n*gger  & n*gger & racial\\ 
小(xiǎo)~仙(xiān)~女(nǔ) & fairy & - & shrew &sexual\\ 
凯(kǎi)~勒(lè)~奇(qí) & Kalergi & - & Kalergi Plan &racial\\ 
\toprule
\end{tabular}  
\vspace{-0.05in}
    \caption{Example illustration of Chinese insults. Among them, \textit{Composition} means the structure and formation of these words, while it is the lexical foundation to derive the \textit{Actual Meaning}. }
\vspace{-0.05in}
\label{token_exp}
\end{table*}

In this section, we further explain the term in Section~\ref{lexicon} and list several insults shown in \tablename~\ref{token_exp}, presenting their morphologies to analyze the literal and flexible derived meanings.



\textbf{Deformation.} Since Chinese characters are pictographs, they will be given meanings containing specific emotions by separating and combining with individual characters \cite{chen2012}. An example is "默" (meaning "\textit{silence}"), whose glyph consists of "黑" (meaning "\textit{black}") and "犬" (meaning "\textit{dog}"), implicitly expressing the distaste for the black community.   

\textbf{Homophonic.} Like English, a new word with a similar pronunciation can be substituted for the original word, resulting in the creation of a different semantics \cite{zhang2011}. For instance, netizens always substitute "满" (meaning "\textit{Manchu}") for "蛮" (meaning "\textit{barbarians}"), both of which are pronounced similarly to "\textit{man}".

\textbf{Irony.} Positive words is sometimes ironically used to achieve the effect of insults, which is often reflected in old words with new meanings \cite{DBLP:journals/csur/FortunaN18}. Like "仙女" (meaning "\textit{fairy}"), what was originally a gentle and kind image is implied to be a rude and impolite "\textit{shrew}".

\textbf{Abbreviation.} Shortening and contracting sensitive words will make expressions more concise and clear \cite{chen2012}. An example is "\textit{txl}", where each letter is the pronounced initials of "同", "性", and "恋", respectively, meaning "\textit{gay}".

\textbf{Metaphor.} Internet users often degrade their attacking targets into something sarcasm, such as animals, in order to insult them \cite{zhang2011}. In the term "蠢驴" (meaning "\textit{silly donkey}"), the publisher compares men to donkeys, transmitting aggression against  others.

\textbf{Code Mixing.} To emphasize the tone, non-Chinese language codes are widely mixed in the text on the Chinese web platforms, such as English and emoji \cite{li2020swearwords}. Like profanity "ni哥" (meaning "\textit{ni brother}"), which has the same pronunciation as "\textit{n*gger}".

\textbf{Borrowed Word.} Certain toxic cultural connotations pervade some phonetic foreign words \cite{shi2010web}. Therefore, background information is required to clarify the actual semantics of these terms. An example is "凯勒奇", a reference to the anti-Semitic \textit{Kalergi Program}, which is used as an inflammatory term.

\section{Supplement of Data Description}\label{Appendix_sta} 
\setcounter{table}{0}  
\begin{table*}[t]
\small
  \centering
    \begin{tabular}{l|l}
    \bottomrule
    \multirow{1}{*}[-3pt]{\textbf{Topic}} & \multirow{1}{*}[-3pt]{\textbf{Keywords}} \\ [4pt]
    \hline
    \multirow{2}[2]{*}[-2pt] {\textbf{Gender}} & \multirow{1}{*}[-3pt]{\textit{性别歧视, 性别偏见, 男权主义, 女权主义, 性别对立, 父权, 家庭主妇}} \\[4pt]
          & \multirow{1}{*}[-2pt]{\textit{Sexism, 
    Gender Bias, Masculinity, Feminism, Gender Dichotomy, Patriarchy, Housewife}} \\[4pt]
    \hline
    \multirow{2}[2]{*}[-2pt]{\textbf{Race}} & \multirow{1}{*}[-3pt]{种族歧视, 人种, 黑种人, 白种人, 混血儿, 少数民族, 血统, 肤色, 亚裔} \\[4pt]
          & \multirow{1}{*}[-2pt]{\textit{Racism, Ethnic, Black Race, White Race, Mixed-Blood, Ethnic Minority, Bloodline, Skin Color, Asian}}  \\[4pt]
    \hline
    \multirow{2}[2]{*}[-2pt]{\textbf{Region}} & \multirow{1}{*}[-3pt]{地域歧视, 非洲, 东南亚, 上海, 北京, 广州, 南方人, 北方人} \\[4pt]
          & \multirow{1}{*}[-2pt]{\textit{Regional Discrimination, Africa, Southeast Asia, Shanghai, Beijing, Guangzhou, Northerners, Southerners}}  \\[4pt]
    \hline
    \multirow{2}[2]{*}[-2pt]{\textbf{LGBTQ}} & \multirow{1}{*}[-3pt]{反同性恋, 异性恋, 同性恋, 女同性恋, 男同性恋, 双性向者, 跨性别者, 酷儿, 性取向} \\[4pt]
          & \multirow{1}{*}[-2pt]{\textit{Anti-LGBTQ, Heterosexuality, Homosexuality, Lesbian, Gay, Bisexual, Transgender, Queer, Sexual Orientation}}  \\[4pt]
    \toprule
    \end{tabular}%
    \vspace{-0.025in}
    \caption{Topic and keywords of crawled data.}
    \vspace{-0.025in}
  \label{Keyword}%
\end{table*}%

\subsection{Keywords and Platforms}
\tablename~\ref{Keyword} lists the keywords of the crawled posts for diverse topics. In the process of data annotation, we note that the distribution of toxic subtype has a significant difference in the samples from the two platforms, as shown in \tablename~\ref{Statistics of platforms}. From the statistics, we can see that more than half of the toxic samples on \textit{Zhihu} have subtle expressions, including implicit hate and reporting. In contrast, users from \textit{Tieba} utilize more direct attacks to insult others, containing general offensive language and explicit hate speech. This inspires us to adapt the methodology appropriately to detect toxic language from different platforms in future work. 

\begin{table*}[t]
\renewcommand{\arraystretch}{1.2}
\small
  \centering
    \begin{tabular}{m{1.5cm}<{\centering}|m{1.1cm}<{\centering}|m{1.1cm}<{\centering}|m{1.1cm}<{\centering}|m{1.1cm}<{\centering}|m{1.1cm}<{\centering}m{1.1cm}<{\centering}m{1.1cm}<{\centering}|m{1.1cm}<{\centering}||m{1.1cm}<{\centering}}
    \bottomrule
    \multirow{2}[4]{*}[5pt]{\textbf{Platform}} & \multirow{2}[4]{*}[5pt]{\textbf{N-Tox.}}  & \multirow{2}[4]{*}[5pt]{\textbf{Tox.}} & \multicolumn{5}{c|}{\textbf{Toxic Category}} & \multirow{2}[4]{*}[5pt]{\textbf{Total}} & \multirow{2}[4]{*}[5pt]{\textbf{Avg. \textit{L}}} \\
\cline{4-8}     &      &       & \textbf{Off.} & \textbf{Hate} & \textbf{H-exp.} & \textbf{H-imp.} & \textbf{H-rep.} &       &         \\ 
    \hline
    \textbf{Zhihu} &3,094    & 3,187      &270       &2,917       &1,088       &1,055       &774       &6,281       &41.75  \\
    \textbf{Tieba} &2,456       &3,274       &546       &2,728       &1,649       & 940      & 139      &5,730       &34.99  \\
    \hline
    \textbf{Total} & 5,550   &6,461    & 816      & 5,645      & 2,737      & 1,995      & 913      & 12,011      & 38.37 \\
    \toprule
    \end{tabular}%
    \vspace{-0.025in}   
    \caption{Statistics of different platforms in ToxiCN.}
    \vspace{-0.025in}

  \label{Statistics of platforms}%
\end{table*}%

\begin{table}
\renewcommand{\arraystretch}{1.2}
\small
  \centering
    \begin{tabular}{m{1.5cm}<{\centering}|m{1.5cm}<{\centering}|m{1.5cm}<{\centering}|m{1.5cm}<{\centering}}
    \bottomrule
    \textbf{Lexicon} & \textbf{Label} & \textbf{w/ Insult} & \textbf{w/o Insult} \\
    \hline
    \multirow{3}[4]{*}[5pt]{\textbf{Pseudo}} & \textbf{Tox.} & 3,166  & 3,295 \\
\cline{2-4}          & \textbf{N-Tox.} & 308   & 5,242 \\
\cline{2-4}          & \textbf{Total} & 3,474   & 8,537 \\
    \hline
    \hline
    \multirow{3}[4]{*}[5pt]{\textbf{Annotation}} & \textbf{Tox.} & 4,331  & 2,130 \\
\cline{2-4}          & \textbf{N-Tox.} & 1,162  & 4,388 \\
\cline{2-4}          & \textbf{Total} & 5,439   & 6,518 \\
    \toprule
    \end{tabular}%
    \vspace{-0.025in}
    \caption{Statistics of samples with ("\textit{w/}") or without ("\textit{w/o}") any insults, where "\textit{Pseudo}" and "\textit{Annotation}" means the insults are from the profanity list in the pseudo labeling and the final lexicon respectively.}
    \vspace{-0.025in}

  \label{if_w_insult}%
\end{table}%

\subsection{Samples \textit{w} or \textit{w/o} insults}

In this section, we statistic the samples containing insults to further demonstrate the effectiveness of the two-step annotation procedure. To restore the process, we count the samples based on the profanity list at the end of the pseudo-annotation phase and the final insult lexicon, respectively. The result is shown in \tablename~\ref{if_w_insult}.

In the first stage, 3,474 samples are pseudo-labeled as toxic, and 91.1\% of them are indeed toxic, representing 49\% of all toxic samples. This reflects the fact that pseudo-annotation can significantly reduce the annotation burden. Afterwards, some low-frequency swear words and implicit insults are added to the lexicon during the main manual labeling. 1,162 instances with insults are ultimately labeled as non-toxic and 2,130 without insults are toxic, showing the necessity of manual inspection. These samples are more difficult to be identified than the cases that can be filtered directly using an insult lexicon, and need to be focused on in future work. However, even so, the comments containing insults are more likely to be toxic, illustrating the significance of lexical knowledge for toxic language detection.        




\subsection{Samples of Attack Multi-class Group}
Here we calculate the proportion of utterances attacking multi-class groups. As the result shown in \tablename~\ref{multi-class group}, there are about 15\% of the samples containing attacks and discrimination against multiple groups in \textsc{ToxiCN}.

\begin{table}
\renewcommand{\arraystretch}{1.2}
\small
  \centering
    \begin{tabular}{m{3.5cm}<{\centering}m{1cm}<{\centering}m{1cm}<{\centering}}
    \bottomrule
    \textbf{Num of Group Labels (n)} & \textbf{Size}  & \textbf{\%} \\
    \hline
    1     & 4,802      & 85.07  \\
    2     & 788      & 13.96  \\
    $\geq3$     & 54      & 0.96 \\
    \toprule
    \end{tabular}%
    \vspace{-0.025in}
\caption{Statistics of hate speech against single and multi-class attacked groups. \textit{Size} denotes the number of samples attacking \textit{n} groups, and \% is the percentage of matched samples of the total number of hate speech.}
    \vspace{-0.025in}
\label{multi-class group}%
\end{table}%

\subsection{Statistics of Sub-varieties of Chinese}

Simplified Chinese accounts for 99.7\% of the crawled data on both platforms. This reflects the reality of Chinese web platforms because Simplified Chinese is the official language of China. And we do not do additional sampling for data imbalance.

\section{Experimental Details}\label{Appendix_exp} 

The details of the hyperparameters are listed in \tablename~\ref{hyp}. During the phase of the experiment, we note that the hyperparameters of BERT and RoBERTa with the best performance are basically the same.

\setcounter{table}{0}  
\begin{table}[htpb]
\renewcommand{\arraystretch}{1.2}
\small
\centering
    \begin{tabular}{m{2.6cm}<{\centering}m{1.2cm}<{\centering}m{2.5cm}<{\centering}}
    \hline
    \textbf{Hyperparameters} & \textbf{BiLSTM} & \textbf{BERT/RoBERTa}\\
    \hline
    epochs &20  &5  \\
    batch size &64  &32  \\
    learning rate &1e-3  & 1e-5 \\
    padding size & 100 & 80\\
    dropout rate &0.5  & 0.5 \\
    $\lambda$ & 0.5 & 0.01 \\
    \hline
    \end{tabular}
    \caption{The hyperparameters of the experiment.}
    \label{hyp}
\end{table}



\clearpage

\clearpage\end{CJK*}
\end{document}